\documentclass{article}
\usepackage{spconf,amsmath,graphicx}
\usepackage{xcolor}
\usepackage{booktabs}
\usepackage{multirow}
\usepackage{siunitx}

\usepackage{times}
\usepackage{epsfig}
\usepackage{graphicx}
\usepackage{amsmath}
\usepackage{amssymb}
\usepackage{subfig}
\usepackage{float}
\usepackage{hyperref}

\title{Multi-Branch with Attention Network for Hand-Based Person Recognition}

\name{
\begin{tabular}{c} Nathanael L. Baisa, Bryan Williams, Hossein Rahmani, Plamen Angelov, Sue Black \end{tabular} 
\thanks{
Nathanael L. Baisa is with the School of Computer Science and Informatics, De Montfort University, Leicester LE1 9BH, UK. Email: nathanael.baisa@dmu.ac.uk. }
\thanks{
Bryan Williams, Hossein Rahmani, Plamen Angelov and Sue Black are with the School of Computing and Communications, Lancaster University, Lancaster LA1  4WA, UK. Emails: \{b.williams6, h.rahmani, p.angelov, sue.black\}@lancaster.ac.uk. }
}
\address{}

\begin{document}
%
\maketitle
\begin{abstract}

In this paper, we propose a novel hand-based person recognition method for the purpose of criminal investigations since the hand image is often the only available information in cases of serious crime such as sexual abuse. Our proposed method, Multi-Branch with Attention Network (MBA-Net), incorporates both channel and spatial attention modules in branches in addition to a global (without attention) branch to capture global structural information for discriminative feature learning. The attention modules focus on the relevant features of the hand image while suppressing the irrelevant backgrounds. In order to overcome the weakness of the attention mechanisms, equivariant to pixel shuffling, we integrate relative positional encodings into the spatial attention module to capture the spatial positions of pixels. Extensive evaluations on two large multi-ethnic and publicly available hand datasets demonstrate that our proposed method achieves state-of-the-art performance, surpassing the existing hand-based identification methods. The source code is available at \textcolor{red}{\url{https://github.com/nathanlem1/MBA-Net}}. 

\end{abstract}
\section{Introduction} \label{sec:intro}

Biometric identification, recognition of individuals using body parts or behavioural characteristics, has recently received a lot of attention for different applications. Hand images, one of the primary biometric traits~\cite{DanEliRos16, BaiJiaVya22}, deliver discriminative features for biometric person recognition. Hand images not only have less variability when compared to other biometric modalities but also have strong and diverse features which remain relatively stable after adulthood~\cite{BaiJiaVya22, YimChaShu20, AttAkhCha21}. Because of this, there is a strong potential to investigate hand images captured by digital cameras for person recognition, especially for criminal investigation in uncontrolled environments since they are often the only available information in cases of serious crime such as sexual abuse.

Both traditional and deep learning approaches have been combined in~\cite{Mah19,YimChaShu20} to develop person identification using hand images. After training a convolutional neural network (CNN) on digital hand images (RGB), the method in~\cite{Mah19} used the network as a feature extractor to obtain CNN-features which have been fed into a set of support vector machine (SVM) classifiers. The work in~\cite{YimChaShu20} used rather similar approach with additional data type for fusion, near-infrared (NIR) images. However, these methods are not an end-to-end. Recently, self-attention mechanism, an integral component of Transformers~\cite{KhaNasHay21}, has received great attention in deep learning. The self-attention mechanism captures long-term information and interactions amongst all entities (e.g. pixels, channels, sequence elements, etc.) of the input data. It updates each pixel, for instance, by aggregating global information from all pixels in the input image. The attention mechanism has been used in~\cite{TiaShaJin19, ZhiCuiWen20} by considering both channel and spatial attentions which compute correlations between all the channels and all the pixels of the input feature map, respectively. However, these methods used the attention modules across the entire network layers which make it computationally inefficient since the self-attention computation is very expensive if the dimension of the input data is very large. Furthermore, these methods are designed for body-based person re-identification and have limited performance when applied to hand images. By default, the attention mechanism does not model relative or absolute position information. To overcome this, a relative position representation of a sequence element with respect to its neighbours has been proposed in~\cite{ShaJakAsh18} for natural language processing, and later on incorporated into a standalone global attention-based deep network for images without using convolutions for modeling pixel interactions~\cite{SheBelVem20}. Unlike these methods, we use the attention mechanism along with the convolution operations by applying the attention modules only to low-resolution feature maps in later stages of a deep network in branches for better computational efficiency as well as accuracy for hand-based person identification.

In this work, we propose a novel hand-based person recognition method, Multi-Branch with Attention Network (MBA-Net), by learning attentive deep feature representations from hand images captured by digital cameras for criminal investigations. The proposed method is trained in an end-to-end manner. Our contributions can be summarized as follows. 

\begin{enumerate}
\item We propose a network by incorporating both channel and spatial attention modules in branches in addition to the global (without attention) branch for hand-based person recognition which is efficient computationaly and flexible in terms of the backbone architecture.
\item We include relative positional encodings into the spatial attention module, considering height and width independently, to capture the spatial positions of pixels in order to overcome the weakness of the attention mechanisms - equivariant to pixel shuffling, for efficiently recognizing suspects based on hand images.
\item We make extensive evaluations on two large multi-ethnic and public hand datasets (11k hands~\cite{Mah19} and HD~\cite{KumXu16}  datasets). 
\end{enumerate}

The rest of the paper is organized as follows. The proposed method including the attention modules and the overall architecture of MBA-Net is described in Section~\ref{sec:proposedMethod} followed the experimental results in Section~\ref{sec:experimentalResults}. The main conclusion along with suggestion for future work is summarized in Section~\ref{sec:Conclusion}.

\section{Proposed Method} \label{sec:proposedMethod}

In this section, we introduce the two attention modules followed by the overall architecture of MBA-Net. The goal of the attention modules is to supress irrelevant backgrounds while focusing on discriminative information of hand appearances.

\subsection{Channel Attention Module}

Channel attention module (CAM) aims to aggregate channel-wise feature-level information since some channels in higher convolutional layers are semantically related i.e. CAM computes the correlations between all the channels. The structure of CAM is given in Fig.~\ref{fig:CAM}. Given the input feature map $\mathbf{E}_i \in \mathbb{R}^{C \times H \times W}$ where C, H, W are the number of channels, height and width of the feature map, respectively, we first reshape it to produce the matrices of keys, queries and values, respectively, as $\mathbf{K} \in \mathbb{R}^{C \times HW}$, $\mathbf{Q} \in \mathbb{R}^{C \times HW}$ and $\mathbf{V} \in \mathbb{R}^{C \times HW}$. Then, the global channel attention map $\mathbf{A}_c \in \mathbb{R}^{C \times C}$ is computed using the dot-product of the query with all keys as

\begin{equation}
    \mathbf{A}_c = \rho (\mathbf{K} \mathbf{Q}^T)
\label{eq:Ac}
\end{equation}
\noindent where $\mathbf{Q}^T$ denotes the matrix transpose of $\mathbf{Q}$, and $\rho$ represents the softmax normalization along each row separately. The self-attended output feature map $\mathbf{E}_o \in \mathbb{R}^{C \times H \times W}$ for the channel attention is given by

\begin{equation}
    \mathbf{E}_o = \gamma (\mathbf{A}_c \mathbf{V}) + \mathbf{E}_i
\label{eq:Ac}
\end{equation}
\noindent where $\gamma$ is initialized as 0 and gradually learns to assign more weight to adjust the impact of the CAM. $\mathbf{A}_c \mathbf{V}$ means that the matrix of values $ \mathbf{V}$ is weighted by the attention score $\mathbf{A}_c$.

\subsection{Spatial Attention Module with Relative Positional Encodings}

\begin{figure*}[htbp]
  \begin{center}
   \subfloat[Channel Attention Module (CAM)]
  {\label{fig:CAM} \includegraphics[width=0.80\linewidth]{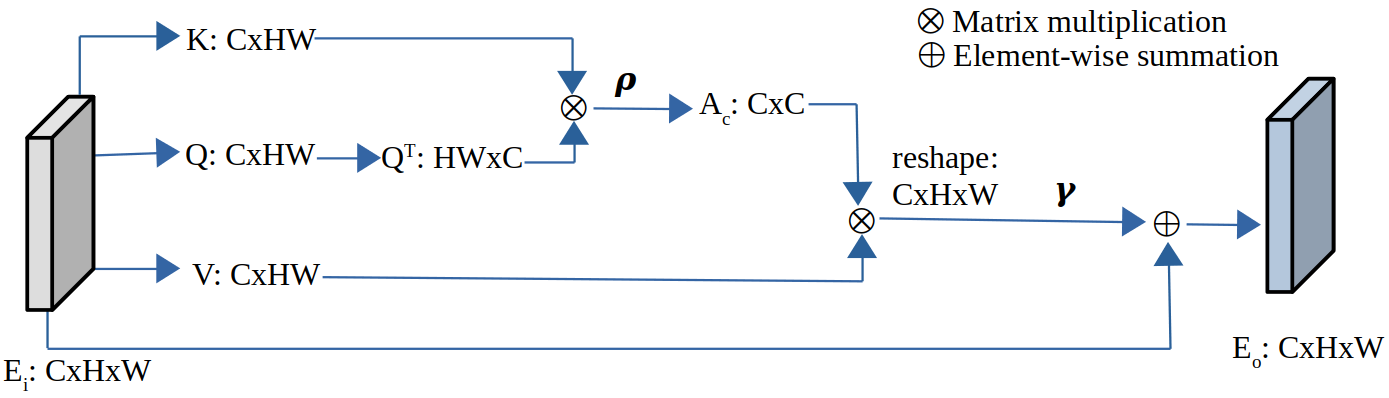}} \\
  \subfloat[Spatial Attention Module with Relative Positional Encodings (SAM-RPE)]
  {\label{fig:SAM-RPE} \includegraphics[width=0.80\linewidth]{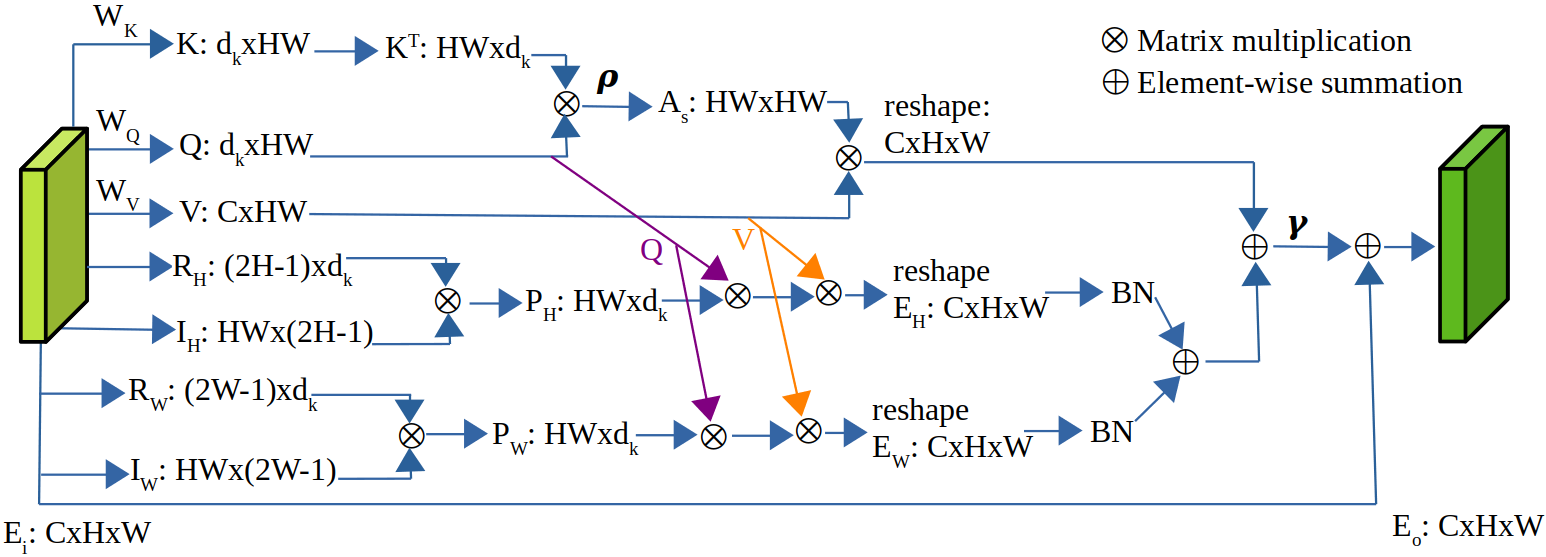}}  
  \end{center}
   \caption{Attention modules, CAM and SAM-RPE, used in our proposed MBA-Net.}
  \label{fig:CAM_SAM-RPE}
\end{figure*}

The goal of spatial attention module is to aggregate the semantically similar pixels in the spatial domain of the input feature map. Though the spatial attention mechanism attends to the entire input feature map based on content (pixel values), it does not take into account the spatial positions of pixels which makes it equivariant to pixel shuffling. To overcome this, we incorporate the relative positional encodings along the rows (height) and columns (width), which is computationally efficient, so that it maintains translation equivariance i.e. translating (shifting) the input pixel also translates the output pixel by the same amount. The structure of the Spatial Attention Module with Relative Positional Encodings (SAM-RPE) is given in Fig.~\ref{fig:SAM-RPE}. 
Given the input feature map $\mathbf{E}_i \in \mathbb{R}^{C \times H \times W}$, the matrices of keys $\mathbf{K} \in \mathbb{R}^{d_k \times HW}$, queries $\mathbf{Q} \in \mathbb{R}^{d_k \times HW}$ and values $\mathbf{V} \in \mathbb{R}^{C \times HW}$ are obtained by transforming it through defined learnable weight matrices $\mathbf{W}_K$, $\mathbf{W}_Q$ and $\mathbf{W}_V$, respectively, where $d_k = \frac{C}{8}$ is the channels dimension of the keys and queries. The learnable weight matrices $\mathbf{W}_K$, $\mathbf{W}_Q$ and $\mathbf{W}_V$ are implemented using independent pointwise ($1 \times 1$) convolution layers with batch normalization and ReLU activation. Thus, the global spatial attention map $\mathbf{A}_s \in \mathbb{R}^{HW \times HW}$ is computed as

\begin{equation}
    \mathbf{A}_s = \rho (\mathbf{K}^T \mathbf{Q})
\label{eq:Ac}
\end{equation}
\noindent where $\mathbf{K}^T$ denotes the matrix transpose of $\mathbf{K}$, and $\rho$ represents the softmax normalization along each row separately.

We consider height (row) and width (column) attentions due to the relative spatial positions for computational efficiency. To compute these relative positional attentions, we first need to represent relative shifts along the height or width of the input feature map. Let a relative position embedding for the height that needs to be learned be $\mathbf{R}_H \in \mathbb{R}^{(2H-1) \times d_k}$ where $H$ is the height of the input feature map and $d_k =\frac{C}{8}$ is the number of channels. A possible vertical shift, from $-(H-1)$ to $H-1$, corresponds to each row of $\mathbf{R}_H$. The relative shifts in the matrix $\mathbf{R}_H$ need to be represented using absolute shifts. To do this, the re-indexing tensor $\mathbf{I}^H \in \mathbb{R}^{H \times W \times (2H-1)}$, which is used as a mask, can be defined as

\begin{equation}
    \mathbf{I}^H_{h,i,r} =
\begin{cases}
    1,  &  \text{if}~ i-h = r~~ \& ~~ |i-h |  \leq H \\
    0,  & \text{otherwise}
\end{cases}
\label{eq:labelSmoothing}
\end{equation}
\noindent where $h \in \{0, ..., H-1\}$, $i \in \{0, ..., W-1\}$ and $r \in \{-(H-1), ..., 0, ..., H-1\}$.

By reshaping $\mathbf{I}^H$ to $\mathbf{I}_H \in \mathbb{R}^{HW \times (2H-1)}$, a position embedding tensor with indices of absolute shifts  for the height $\mathbf{P}_H \in \mathbb{R}^{HW \times d_k}$ is given by 

\begin{equation}
    \mathbf{P}_H = \mathbf{I}_H \mathbf{R}_H
\label{eq:Ph}
\end{equation}
\noindent Then, the self-attended feature map $\mathbf{E}_H \in \mathbb{R}^{C \times H \times W}$ corresponding to the height relative position embedding $\mathbf{R}_H$, which is used as keys implicity ($\mathbf{P}_H$ explicitly), is computed as 

\begin{equation}
    \mathbf{E}_H = \mathbf{V} (\mathbf{P}_H \mathbf{Q})
\label{eq:Ph}
\end{equation}
\noindent where $\mathbf{P}_H \mathbf{Q}$ corresponds to the height relative positional attention.

The relative position embedding for the width $\mathbf{R}_W \in \mathbb{R}^{(2W-1) \times d_k}$, the re-indexing tensor $\mathbf{I}_W \in \mathbb{R}^{HW \times (2W-1)}$ and its corresponding self-attended feature map $\mathbf{E}_W \in \mathbb{R}^{C \times H \times W}$  can be obtained with similar approach to the above height formulation since they are symmetric.

Thus, the final self-attended output feature map $\mathbf{E}_o \in \mathbb{R}^{C \times H \times W}$ for the SAM-RPE is given by

\begin{equation}
    \mathbf{E}_o = \gamma (\mathbf{V} \mathbf{A}_s + BN(\mathbf{E}_H) + BN(\mathbf{E}_W)) + \mathbf{E}_i
\label{eq:Ac}
\end{equation}
\noindent where $\gamma$ is initialized as 0 and gradually learns to assign more weight to adjust the impact of the SAM-RPE, and $BN$ is a batch normalization.


\noindent

\subsection{Network Architecture Overview}

\begin{figure*}[t]
\begin{center}
  \includegraphics[width=0.8\linewidth]{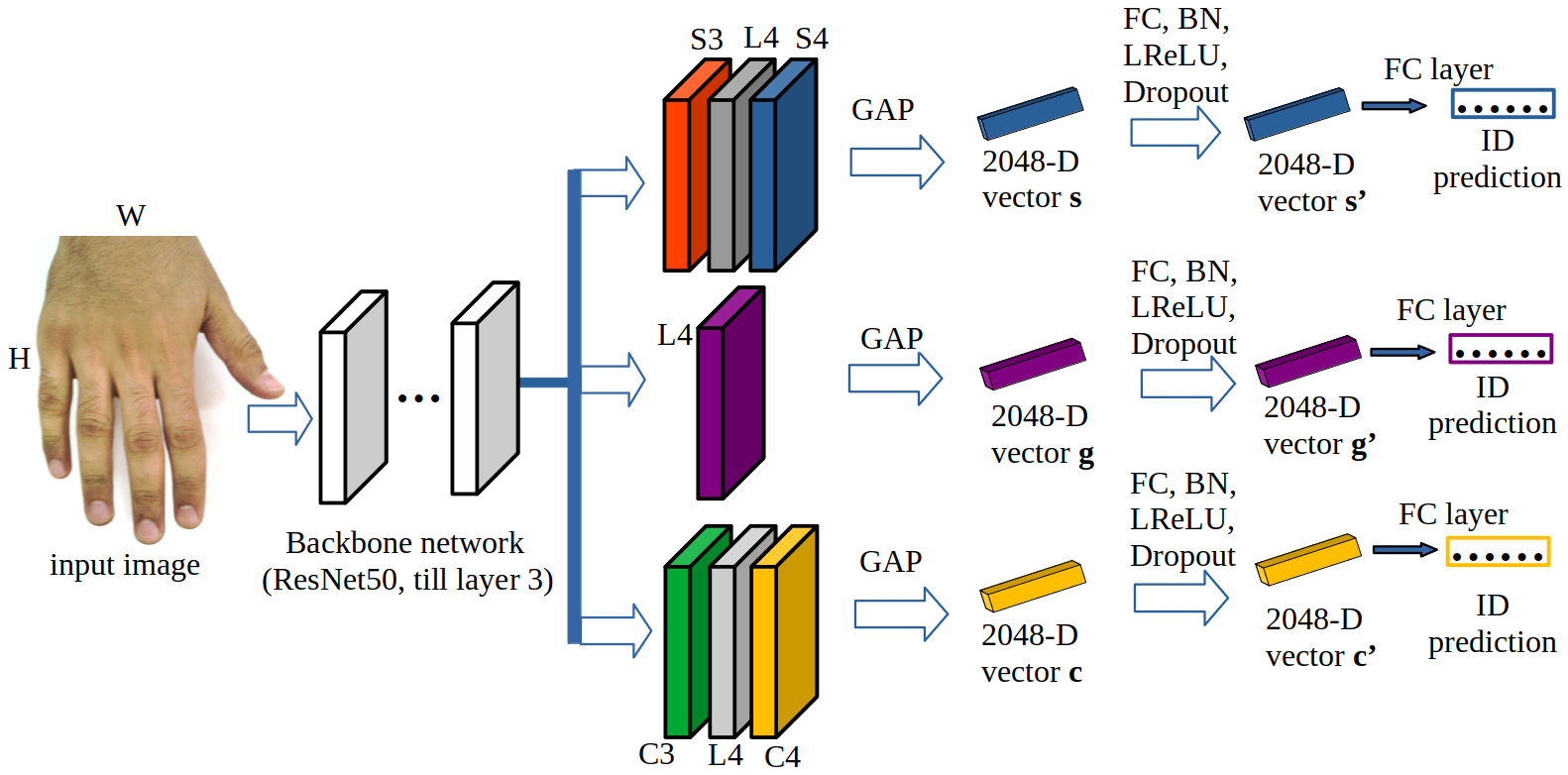} \\
\end{center}
   \caption{Structure of MBA-Net. S3 and S4 are the SAM-RPE after layer 3 and layer 4 (L4) of the ResNet50, respectively. Similarly, C3 and C4 are the CAM after layer 3 and layer 4 of the ResNet50, respectively. Given an input image, three separate 2048-D column feature vectors are obtained by passing it through the backbone network with the 3 branches. Each classifier predicts the identity of the input image during training.}
\label{fig:MBAdiagram}
\end{figure*}

The overall proposed method, MBA-Net, is given in Fig.~\ref{fig:MBAdiagram}. It incorporates two complementary attention modules, channel and spatial. These attention modules are used at higher level of the network, for computational efficiency, in branches along with the global without attention branch. As a backbone network, we use ResNet50~\cite{HeZhaSun16} pretrained on ImageNet due to its precise architecture with competitive performance. Obviously, any network designed for image classification can be adapted, for example Inception network~\cite{ChrSerVin17}. We keep the structure of the original ResNet50 before layer 3 (inclusive) remain the same when we modify the backbone network to produce the MBA-Net. We create 3 branches just after the layer 3 of the ResNet50 to incorporate the channel and the spatial (with relative positional encodings) attention modules in branches by keeping one branch for the without attention global classifier.   

For each branch, a new fully-connected layer (FC), batch normalization (BN), leaky rectified lineat unit (LReLU) and dropout with probability of 0.5 to reduce possible over-fitting are employed to process 2048-dimensional column feature vectors obtained after the global average pooling (GAP) which in turn are fed into a classification layer. The classification layer, which is implemented using a FC layer followed by a softmax function, predicts the identity (ID) of each input. 
In addition, we change the last stride from 2 to 1 in the backbone network i.e. remove the last spatial down-sampling operation, which increases the size of the tensor of each branch for improved performance as observed in~\cite{LuoGuLia19}.

The MBA-Net is optimized during training by minimizing the sum of the Cross-Entropy losses over the 3 ID predictions i.e. each classifier predicts the identity of the input image. During testing, we concatenate all the 2048-D feature vectors of the 3 branches, just after the GAP i.e. $\mathcal{F} = [\textbf{s}, \textbf{g}, \textbf{c}]$ which becomes 6144-D feature vector, and then compare feature vector of each query image with gallery feature vectors using cosine distance.

\noindent

\section{Experimental Results}  \label{sec:experimentalResults}

\subsection{Settings}

\begin{figure}[htbp] 
\begin{center}
  \includegraphics[width=1.0\linewidth]{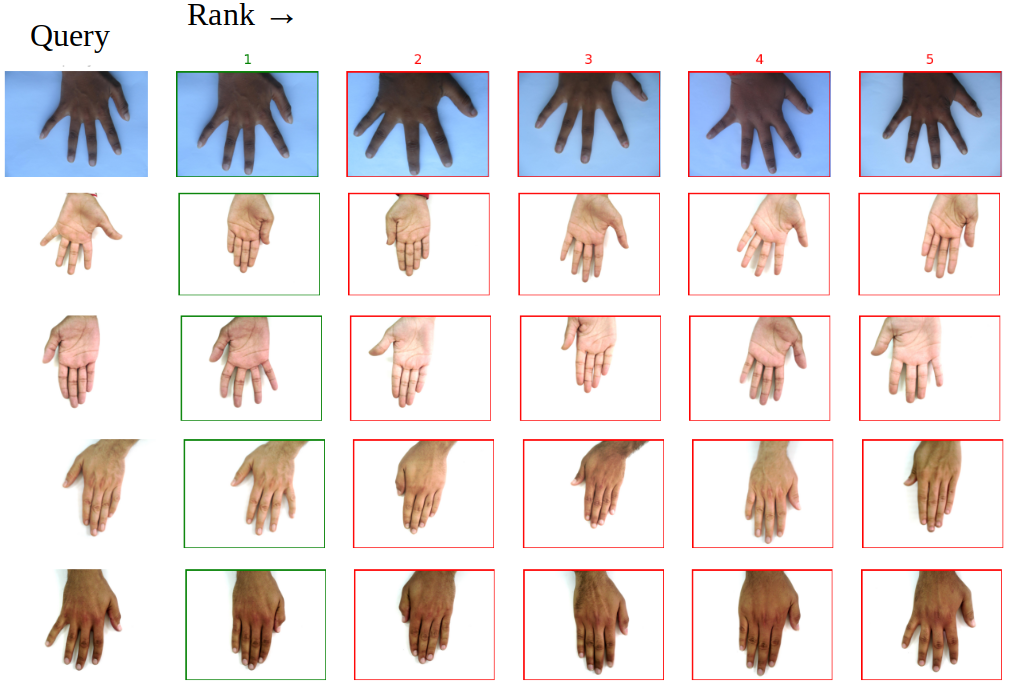} \\
\end{center}
   \caption{Some qualitative results of our method using query vs ranked results retrieved from gallery. From top to bottom row are HD, left palmar of 11k, right palmar of 11k, left dorsal of 11k and right dorsal of 11k datasets. The green and red bounding boxes denote the correct and the wrong matches, respectively.}
\label{fig:demoResult}
\end{figure}

\textbf{Datasets:} For evaluation, we use two different datasets, 11k hands\footnote{\url{https://sites.google.com/view/11khands}} dataset~\cite{Mah19} and Hong Kong Polytechnic University Hand Dorsal (HD)\footnote{\url{http://www4.comp.polyu.edu.hk/~csajaykr/knuckleV2.htm}} dataset~\cite{KumXu16}. The former has 190 subjects (identities) whereas the latter has 502 identities. We use the same partitioning strategy of the datasets as in~\cite{BaiJiaVya22}. As in~\cite{BaiJiaVya22}, the 11k hands dataset is divided into right dorsal, left dorsal, right palmar and left palmar sub-datasets to train a hand-based person recognition model. After excluding accessories and dividing the dataset as in~\cite{BaiJiaVya22},  right dorsal has 143 identities, left dorsal has 146, right palmar has 143 and left palmar has 151 identities. For all 11k sub-datasets and HD dataset, the first half and the second half of each (sub-)dataset based on identity are used for training and testing, respectively. For instance, for right dorsal, the first 72 identities are used for training and the last 71 identities are used for testing. Similarly, the first 73, 72, 76 and 251 identities are used for training phase for left dorsal, right palmar, left palmar and HD datasets, respectively. The remaining identities of each (sub-)dataset (73 for left dorsal, 71 for right palmar, 75 for left palmar, 251 for HD) are used for testing. From each identity of the test set of each 11k sub-dataset, we randomly choose one image and put in a common gallery for all the 11k sub-datasets. The remaining images of each identity of the test set of each sub-dataset are used as a query set for that sub-dataset. Similarly, from each identity of the test set of the HD dataset, one image is randomly chosen and put in a gallery and the rest are used as a query (probe). Unlike the 11k dataset, the HD dataset has additional images of 213 subjects, which lack clarity or do not have second minor knuckle patterns, and are added to the HD gallery. Accordingly, the 11k gallery has 290 images and the query has 971 images for right dorsal, 988 images for left dorsal, 917 images for right palmar and 948 images for left palmar. Similarly, the gallery for the HD dataset has 1593 images and the query has 1992 images. A randomly chosen image of each identity of the training set of each dataset is used as a validation for monitoring the training process. This procedure is repeated for 10 times and the average performance is reported. We show a qualitative exemplar image (query) of each (sub-)dataset in Fig.~\ref{fig:demoResult} with ranked results retrieved from a gallery of each (sub-)dataset.

\noindent \textbf{Implementation details:} We implemented the MBA-Net using PyTorch deep learning framework and trained it on NVIDIA GeForce RTX 2080 Ti GPU. The input images are resized to $356 \times 356$ and then randomly cropped to $324 \times 324$, augmented by random horizontal flip, normalization and color jittering during training. To prevent over-fitting and over-confidence, label smoothing~\cite{SzeVanIof16} with smoothing value ($\epsilon$) of 0.1 is also used. However, only normalization is utilized during testing with the test images resized to $324 \times 324$, without a random crop. In both cases, a random order of images are used by reshuffling the dataset. We train the model for 70 epochs with mini-batch size of 20 using Cross-Entropy loss and Adam optimizer with the weight decay factor for L2 regulization of $5 \times 10^{-4}$. For the first 10 epochs, we use a warmup strategy~\cite{LuoGuLia19}, increasing a learning rate linearly from $8 \times 10^{-6}$ to $8 \times 10^{-4}$, and then it is decayed to $4 \times 10^{-4}$, $2 \times 10^{-4}$ after 40 and 60 epochs, respectively. The learning rate is divided by 10 for the existing layers of the backbone network i.e. ten times bigger learning rate is given to the newly added layers (FC layers and batch normalizations) and the attention modules (embedding functions and batch normalizations), with appropriate weight and bias initializations.

\subsection{Performance Evaluation}

We evaluate our model and report the results using mean Average Precision (mAP) and Rank-1 matching accuracy~\cite{ZheSheTia15} on 11k~\cite{Mah19} and HD~\cite{KumXu16} datasets. We compare our proposed method, MBA-Net, to many other state-of-the-art methods such as GPA-Net~\cite{BaiJiaVya22}, RGA-Net~\cite{ZhiCuiWen20} and ABD-Net~\cite{TiaShaJin19}. The GPA-Net was designed for hand-based person identification, however, both RGA-Net and ABD-Net were designed for body-based person re-identification. Therefore, we trained both RGA-Net and ABD-Net on hand datasets using the same experimental settings (loss function, optimizer, hyperparameters, etc.) as our method to make a fair comparasion with our method. The quantitative same-domain performance comparison of our method with the other methods is given in Table~\ref{tbl:Comparison}. The same-domain perforamnce is when the model is trained on a specific dataset and then tested on the test set of that dataset. As shown in this table, our method outperforms all other methods across all datasets in both rank-1 accuracy and mAP evaluation metrics.

\begin{table*} [htbp]
\begin{center}
  \begin{tabular}{|l|ll|ll|ll|ll|ll|}
    \hline
    \multirow{2}{*}{Method} & 
      \multicolumn{2}{c|}{D-r of 11k} & 
      \multicolumn{2}{c|}{D-l of 11k} & 
      \multicolumn{2}{c|}{P-r of 11k} &
      \multicolumn{2}{c|}{P-l of 11k} &
      \multicolumn{2}{c|}{HD} \\
      \cline{2-11}
    & rank-1 & mAP & rank-1 & mAP & rank-1 & mAP & rank-1 & mAP & rank-1 & mAP \\
    \hline
	GPA-Net~\cite{BaiJiaVya22} & 94.80 & 95.72 & 94.87 & 95.93 & 95.83 & 96.31 & 95.72 & 96.20 & 94.64 & 95.08 \\ 
	RGA-Net~\cite{ZhiCuiWen20} & 94.77 & 95.67 & 95.30 & 95.98 & 92.66 & 93.58 & 94.95 & 95.67 & 95.06 & 95.39 \\ 
	ABD-Net~\cite{TiaShaJin19} & 95.89 & 96.76 & 94.26 & 95.34 & 96.21 & 96.91 & 95.54 & 96.01 & 94.93 & 95.38 \\
	\textbf{MBA-Net (Ours)} & \textbf{97.45} & \textbf{97.98} & \textbf{96.71} & \textbf{97.41} & \textbf{98.05} & \textbf{98.42} & \textbf{97.42} & \textbf{97.84} & \textbf{95.12} & \textbf{95.54}\\
	\hline 
  \end{tabular}
\end{center}
\caption{Same-domain performance comparison of our method (MBA-Net) with other methods (GPA-Net~\cite{BaiJiaVya22}, RGA-Net~\cite{ZhiCuiWen20} and ABD-Net~\cite{TiaShaJin19}) on right dorsal (D-r) of 11k, left dorsal (D-l) of 11k, right palmar (P-r) of 11k, left palmar (P-l) of 11k and HD datasets. The results are shown in rank-1  accuracy (\%) and mAP (\%).}
\label{tbl:Comparison}
\end{table*}
\noindent

The comparison of same-doman and cross-domain performance of a model trained on right dorsal (D-r) of 11k dataset is given in Table~\ref{tbl:crossDomain}. The cross-domain performance is when the model is trained on the training set of one dataset and then evaluated on the test set of the other dataset. Accordingly, for the cross-domain performance, we use the HD test set. As shown in Table~\ref{tbl:crossDomain}, our proposed method outperforms all the other methods in both same-domain and cross-domain performance settings using both evaluation metrics which indicates that our method is more generalizable than the other methods. Though the adapted body-based person re-identification methods, RGA-Net~\cite{ZhiCuiWen20} and ABD-Net~\cite{TiaShaJin19}, have comparable performance as the GPA-Net~\cite{BaiJiaVya22} on the same-domain performance (D-r (11k) $\rightarrow$ D-r (11k) test), they have much lower cross-domain performance (D-r (11k) $\rightarrow$ HD test). For instance, the ABD-Net has the same-domain performance of 95.89\% rank-1 and 96.76\% mAP, however, its cross-domain performance drops very significantly to 71.22\% rank-1 and 73.86\% mAP. Even though the domain shift of the data affects generally all the methods, our proposed method has much more generalizability than all the other methods, 84.41\% rank-1 and 85.98\% mAP, an increase of 13.19\% rank-1 accuracy and 12.12\% mAP over the ABD-Net, for instance.

\begin{table*}[htbp]
\begin{center}
  \begin{tabular}{|l|ll|ll|}
    \hline
    \multirow{2}{*}{Method} & 
      \multicolumn{2}{c|}{D-r (11k) $\rightarrow$ D-r (11k) test} & 
      \multicolumn{2}{c|}{D-r (11k) $\rightarrow$ HD test} \\
      \cline{2-5}
 & rank-1 & mAP & rank-1 & mAP  \\
\hline
GPA-Net~\cite{BaiJiaVya22} & 94.80 & 95.72 & 83.02 & 84.65  \\ 
RGA-Net~\cite{ZhiCuiWen20} & 94.77 & 95.67 & 77.80 & 80.07  \\ 
ABD-Net~\cite{TiaShaJin19} & 95.89 & 96.76 & 71.22 & 73.86   \\ 
\textbf{MBA-Net (Ours)} & \textbf{97.45} & \textbf{97.98}  & \textbf{84.41} & \textbf{85.98} \\
\hline 
\end{tabular} 
\end{center}
\caption{Comparison of same-doman and cross-domain performance of a model trained on right dorsal (D-r) of 11k dataset and then evaluated on test set of the right dorsal of 11k and HD datasets. Rank-1 accuracy (\%) and mAP (\%) are shown.}
\label{tbl:crossDomain}
\end{table*}
\noindent

\subsection{Ablation Analysis}

As described in Section~\ref{sec:proposedMethod}, our proposed method incorporates two complementary attention modules, channel and spatial with relative position, at a higher level of the network in branches along with the global without attention branch. The ablation analysis of these components is given in Table~\ref{tbl:Ablation} with evaluation on right palmar (P-r) of 11k hands dataset. As shown in this table, each component contributes to a performance gain. The global branch (ResNet50 with some modifications) gives rank-1 and mAP of 95.43\% and 95.95\%, respectively. Incorporating the spatial attention without the relative position attention boosts the performance to 96.50\% rank-1 and 96.73\% mAP. Rank-1 accuracy and mAP of 97.41\% and 97.65\%, respectively, are obtained by integrating the channel attention module. Incorporating the relative positional encodings into the spatial attention contributes to the performance gain as well, giving the overall MBA-Net performance of 98.05\% rank-1 and 98.42\% mAP on the right palmar (P-r) of 11k hands dataset as shown on the Table~\ref{tbl:Ablation}.

\begin{table}[htbp]
\begin{center}
\begin{tabular}{|c|c|c|}
\hline
    & rank-1 (\%) & mAP (\%) \\ 
\hline
Global (ResNet50) & 95.43 & 95.95 \\ 
+ Spatial attention & 96.50 & 96.73 \\ 
+ Channel attention & 97.41 & 97.65 \\
+ Relative position & \textbf{98.05} & \textbf{98.42} \\
\hline 
\end{tabular} 
\end{center}
\caption{Ablation analysis on components of MBA-Net on right palmar (P-r) of 11k hands dataset.}
\label{tbl:Ablation}
\end{table}
\noindent

\noindent

\section{Conclusion}  \label{sec:Conclusion}

In this work, we introduce a Multi-Branch with Attention Network (MBA-Net) to learn attentive deep feature representations for person recognition based on the hand images. The MBA-Net incorporates both channel and spatial attention modules in branches in addition to the global (without attention) branch which help to focus on the relevant features of the hand image while supressing the irrelevant backgrounds. We also integrate relative positional encodings into the spatial attention module to capture the spatial positions of pixels to overcome the weakness of the attention mechanisms, equivariant to pixel shuffling. The incorporation of these attention modules in branches as well as keeping the global non-attentive branch separately allows a deeper study of the features of the hand image in less controlled environments for robust recognition of the perpetrators of serious crime. The experimental results on two public hand datasets demonstrate the superiority of the proposed method over the existing state-of-the-art methods.

\bibliographystyle{IEEEbib}
\bibliography{refs}

\end{document}